\documentclass[12pt]{article}

\usepackage[english]{babel}
\usepackage[utf8]{inputenc} % changed from utf8x
\usepackage{newtxtext, newtxmath} % replaces \usepackage{times} for better TNR support
\usepackage{csquotes}

\usepackage{amsmath}
\usepackage{graphicx}
\usepackage{etoolbox}
\usepackage{changepage}
\usepackage{titlesec}
\usepackage[parfill]{parskip}
\usepackage[margin=1in]{geometry}
\usepackage{float}
\usepackage{lipsum} % Package to generate dummy text throughout this template. Remove for real use.
\usepackage{xcolor}
\usepackage{comment}
\usepackage{nameref}

\usepackage{booktabs, multirow} % for borders and merged ranges
\usepackage{soul}% for underlines
\usepackage{xcolor,colortbl} % for cell colors
\usepackage{changepage,threeparttable} % for wide tables

\usepackage{enumitem}

\usepackage{xstring}    % provides \MakeLowercase
\usepackage[style=ieee]{biblatex}
\title{\large \textbf{Human-AI Co-Mentorship in Project-Based Learning:\\A Case Study in Financial Forecasting}} % using \large makes the title approximately 14 pt.
\author{\normalsize Freyaa Chawla\\
\normalsize freyaachawla19@gmail.com\\
\normalsize Archbishop Mitty High School, San Jose, CA
\and
\normalsize Ahan Chawla\\
\normalsize achawla2@nd.edu\\
\normalsize University of Notre Dame, Notre Dame, IN
\and
\normalsize Rishi Singh\\
\normalsize rishi.2028i@iimbg.ac.in\\
\normalsize Indian Institute of Management Bodh Gaya, Bodh Gaya, India
\and
\normalsize Joe Germino\\
\normalsize jgermino@alumni.nd.edu\\
\normalsize University of Notre Dame, Notre Dame, IN
\and
\normalsize Grigorii Khvatskii\\
\normalsize gkhvatsk@nd.edu\\
\normalsize University of Notre Dame, Notre Dame, IN}
\date{} % This leaves the date blank.

\makeatletter % This gets the margins for the title set.
\patchcmd{\@maketitle}{\begin{center}}{\begin{adjustwidth}{0.5in}{0.5in}\begin{center}}{}{}
\patchcmd{\@maketitle}{\end{center}}{\end{center}\end{adjustwidth}}{}{}
\makeatother

\begin{document}
\raggedright
\maketitle
\thispagestyle{empty}
\pagestyle{empty}

%----------------------------------------------------------------------------------------
%  PAPER CONTENTS
%----------------------------------------------------------------------------------------

%\begin{comment}
\section*{Abstract}
This paper reflects on a AI research project carried out by a team of high‑school and early‑undergraduate students under the mentorship of graduate researchers and ably assisted by AI tools. We share our experience in not only on the learning experience for the high school students, but also on how AI tools accelerated the process that enabled the high school students to focus on higher order problem formulation and solution. Although the participants entered the project with limited background in both AI and finance, they showed strong enthusiasm for technical market analysis and ETF price prediction. Traditional learning settings would first teach the necessary methods in a classroom setting and only later let students apply them. In contrast, our project emphasized workflow design: students identified the sequence of steps needed to address the problem and then used AI‑driven tools to execute each step.

During the project the team tackled the challenge of forecasting ETF price movements from historical news data. They began by constructing a web scraper that collected the full text of financial articles together with key metadata (title, URL, and publication date). The scraped articles were subsequently processed with large language models \texttt{gpt‑5‑mini}, which assigned sentiment scores on a –10 to +10 scale and provided a brief justification for each rating. These sentiment time series were used to train a number of forecasting models, including both classical statistical approaches and deep‑learning architectures such as LSTM networks.

We note that the high school students developed the necessary code through iterating with the AI tools, and we used our daily stand-ups to debug and answer conceptual questions. Each of the student was able to dig deeper into their area of interest whether computer science or finance, while collaboratively making a significant advance over the summer of 2025. This project was an important pedagogical exercise on how AI tools can be used for mentoring high school students, allowing them to focus on their specific interests and using the daily stand-ups to focus on problem definition and conceptual understanding. Despite their limited technical qualifications, the students were able to leverage AI tools to build meaningful models with real-world application.
%\end{comment}
%------------------------------------------------

\section{Introduction}

Generative artificial intelligence (AI) tools based on large language models (LLMs) are playing an important role in shaping education and training. Yet they create an apparent tension centered on how and when to use these AI tools within a classroom setting. The use of AI introduces challenges that were not previously encountered by either educators or students. In this sense, AI use can be described as a "wicked problem" \cite{corbin2025wicked} that is difficult to conceptualize or address. One key challenge for educators is finding an appropriate balance between the use of AI tools and more traditional approaches to teaching.

In this work, we address this challenge by exploring what happens when AI is not merely a supplementary tool but a {\it{co-mentor}}, that is, actively guiding students alongside human advisors throughout an extended research project.  We reflect on a ten-week project conducted by a team of two high school students (with one entering college over the course of the project) and early undergraduate students, mentored by a graduate student and a postdoctoral scholar in computer science at the University of Notre Dame, a US-based educational institution. Throughout the project, students were encouraged to use AI tools not only to complete tasks but to learn the concepts and skills required to complete them. 

The project's technical goal, building an AI system to forecast ETF prices using historical price data and news sentiment, provided a rigorous test bed for this pedagogical approach. Because the participating students had limited background in programming and finance, the project demanded substantial skill acquisition. In a traditional learning environment, we would have begun with classroom instruction to teach the requisite theories and techniques before students worked independently to apply this knowledge. Instead, we adopted a workflow-driven approach: under the scholars' supervision, students identified the steps required to complete the project and used AI tools to execute and learn from those steps in real time.  

This approach structured our ten weeks as a series of daily stand-ups (around 30 minutes each) focused on problems to be solved and tasks to be accomplished, with students pursuing specific learning exercises, including doing so through dialogues with the AI tools, between sessions. Through this human-machine co-mentorship model, we observed both accelerated progress and instructive failures that shaped our understanding of AI's role in education. 

%This became a fascinating exercise where we focused our in-person discussions on problems to be solved, analysis to be conducted, and tasks to be accomplished, and then the students followed through on specific learning exercises in their own time and also interacting with AI tools for just-in-time learning. We also used the daily stand-ups  to debug code together, explore the prompts, and continued to build on that learning exercise. Through this human-machine co-mentorship approach, we could really accelerate data collection, model building, and analysis, learning and accomplishing way beyond what anyone of us expected. 

%While the project's original intent was to understand how we can leverage LLMs for sentiment analysis and prediction on ETF movements, it also quickly became an opportunity to understand how LLMs can also become a co-mentor for the high school and undergraduate students. Throughout the project, these students were allowed and encouraged to use AI tools to help them complete their assigned tasks. 
This dual focus, on both the learning process and technical outcomes, motivates the following research questions (RQs). 

\begin{enumerate}[label=\textbf{RQ\arabic*}, align=left, labelsep=1em, leftmargin=1in]
    \item{How does AI-augmented mentorship influence the learning process compared to traditional methods of instruction?}
    \item{What are the most frequent failure modes encountered when using AI in a project-based learning context?}
    \item{Do LLM‑generated sentiment scores improve the predictive performance of ETF price forecasting models?}
\end{enumerate}

Our observations suggest human-AI co-mentorship model accelerated both development and learning, allowing students to engage with higher-level concepts rather than getting mired in implementation details. At the same time, we observed recurring failure modes: LLMs often produced subtle errors in code or analysis that were difficult to detect without domain expertise. These failures, however, became valuable learning opportunities, and students gradually developed the judgment to identify and correct such errors independently.

The rest of the paper is organized as follows: \textbf{\nameref{sec:rw}} reviews prior work on AI‑assisted project‑based learning and ETF forecasting techniques. After that, \textbf{Section \nameref{sec:amr}} details the data collection pipeline, sentiment extraction, model architecture (regression plus classification), and presents performance results in tables. Next, \textbf{ \nameref{sec:reflections}} reflects on how generative AI supported programming, writing, and analysis while highlighting both benefits and pitfalls. After that, \textbf{\nameref{sec:disc}} discusses pedagogical implications of our findings and outlines future research directions. Finally, \textbf{\nameref{sec:lim}} addresses study limitations and suggests avenues for broader validation. \textbf{Please note} that unless specified on AI as a mentor or assistant, the use of the word mentor will imply human mentor, which in this case are the graduate and postdoctoral scholars. 

\section{Related Work}
\label{sec:rw}

We review how our study relates to prior work regarding our two main components: (1) the use of AI in project-based education and (2) the methods used for ETF price modeling. 

\subsection{AI in Project-Based Education}

AI for project based learning was widely explored in the literature. Findings reveal that LLMs tend to underperform human experts in tasks that closely resemble grading when benchmarked \cite{wu2025towards}. At the same time, other studies demonstrate that LLMs can outperform humans in other tasks, such as when generating feedback, as evidenced by \cite{sasaki2025evaluation} and further validated in \cite{fung2024automatic} for even stricter and more sensitive settings such as K-12 learning. These results suggest that while LLMs may lag in some domains, they hold promise for use in broader educational contexts.

Additionally, research has explored how LLM-based platforms can support project‑based learning. One study showed that platform‑integrated LLMs can act as autonomous tutors, guiding students through self‑directed projects and yielding better educational outcomes \cite{zhu2025autopbl}. Additional evidence indicates that LLMs can function as mentors in traditional classroom contexts, such as teaching software architecture, thereby extending their utility beyond informal learning environments \cite{gurtl2024design}.

Studies on the practical adoption of AI tools argue that widespread student use of generative AI necessitates early curriculum integration. Researchers noted that "guerrilla" usage of AI by students is expected to grow and recommended embedding AI from the outset of courses \cite{airecommendations}. Following similar recommendations, frameworks for integrating AI into project‑based learning scenarios have been proposed \cite{AIIEEE}. Real-world practice of integrating AI into educational contexts confirms that following such recommendations and embracing AI increases course satisfaction in traditional academic settings \cite{10.1145/3649217.3653584}, enhances student satisfaction and overall learning outcomes in project‑based courses \cite{zhang2025integrating}, and demonstrates the usability of LLMs even in sensitive contexts such as middle‑school projects \cite{zha2024designingchildcentricailearning}.

Collectively, these studies illustrate both the potential benefits and challenges of incorporating AI into project‑based learning environments. At the same time, no deep case study of AI use for project based learning and its possible failure modes was conducted, highlighting the significance and novelty of our study.

\subsection{ETF Price Modeling}

ETF price modeling has become a well‑studied area of data science.
Over the years a wide range of methods and models have been applied to this problem. Classical machine‑learning approaches such as SVM, KNN, and Naive Bayes have appeared in studies since at least the early 2010s \cite{ucar2013two} and continue to surface in more recent work \cite{piovezan2024machine}. Traditional time‑series models (e.g. ARIMA) remain popular for financial data \cite{sharma2025multimodel}. More advanced tree‑based algorithms such as XGBoost \cite{li2026xgboost} also find use in the domain. Finally, deep‑learning techniques have been extensively explored for ETF price prediction \cite{kundu2025predicting, 10.1145/3635638.3635640}.

Unlike many investigations that aim to construct automated trading robots, our focus is not on building a production system. Instead, we intend to guide students through a real‑world project that employs AI for ETF price modeling and observe how they apply the techniques in practice. Having situated our work within the literatures on AI-assisted education and ETF forecasting, we now turn to the methods employed in our study. In describing these methods, we attend not only to technical details but also to how each phase of the project created opportunities for student learning.

\section{Data, Methods and Analysis}
\label{sec:amr}

We describe the methods we used for ETF price modeling, as well as the results and performance metrics of the resulting models. All this work was led by the high school students with support from the undergraduate student (specifically on the price modeling work), and mentorship by the scholars. Importantly, we designed each phase of the pipeline not only to produce functional outputs but also to create structured learning opportunities. Students encountered each technical challenge first through AI-assisted exploration, then refined their understanding through scholar mentor-guided debugging and discussion. 

\subsection{Web Scraping and Sentiment Modeling}

For this study we compiled a historical price dataset for 29 exchange‑traded funds (ETFs), covering January 2, 2024 through September 23, 2025. The data were harvested via Yahoo Finance API \cite{aroussi_ranaroussi/yfinance_2026}, from which we have downloaded daily closing prices for the ETFs of interest. The ETFs under study were chosen because they collectively span the major sectors of the U.S. equity market, including, but not limited to technology, energy, banking \& financials and healthcare. Each ETF was assigned to its primary sector by examining the underlying holdings list provided by the fund sponsor; for example, XLF was mapped to financials, while XOP fell under oil and gas.

\begin{table}[!htp]\centering
\begin{tabular}{lccc|lcccc}\toprule
\textbf{ETF} &\textbf{\#Sentiment} &\textbf{\#Price} &\textbf{Coverage} &\textbf{ETF} &\textbf{\#Sentiment} &\textbf{\#Price} &\textbf{Coverage} \\\midrule
BBJP &55 &436 &12.61\% &SPY &55 &436 &12.61\% \\
CLOU &0 &433 &0.00\% &SRVR &25 &434 &5.76\% \\
DXJ &55 &436 &12.61\% &VDE &104 &449 &23.16\% \\
EUFN &55 &436 &12.61\% &VFH &55 &436 &12.61\% \\
EWJ &55 &436 &12.61\% &VGK &55 &436 &12.61\% \\
EZU &55 &436 &12.61\% &VGT &0 &433 &0.00\% \\
FEZ &55 &436 &12.61\% &VHT &1 &433 &0.23\% \\
FLJP &55 &436 &12.61\% &VOO &55 &436 &12.61\% \\
IEUR &55 &436 &12.61\% &VTI &55 &436 &12.61\% \\
IVLU &55 &436 &12.61\% &XLE &104 &449 &23.16\% \\
IVV &55 &436 &12.61\% &XLF &55 &436 &12.61\% \\
IXJ &1 &433 &0.23\% &XLRE &25 &434 &5.76\% \\
KRE &55 &436 &12.61\% &XLV &1 &433 &0.23\% \\
QQQ &0 &433 &0.00\% &XOP &104 &449 &23.16\% \\
REZ &25 &434 &5.76\% & & & & \\
\bottomrule
\end{tabular}
\caption{Sentiment Coverage by ETF}\label{tab:sentcov}
\end{table}

This data collection phase served as an introduction to working with APIs and structured financial data. Students used AI tools to generate initial API query code, which worked for the basic requests but required mentor intervention to handle rate limiting and error recovery, which served as an early lesson in the gap between AI-generated scaffolding and production-ready code. 

To enrich this pricing dataset with market‑sentiment information, we scraped news stories relevant to the same set of ETFs from the NASDAQ website. We first automated the website's internal search API to download the metadata (publication dates, titles, and links to the full text) of the news articles related to each of the ETFs under analysis. Next, we downloaded the full text of each of the news articles. In total, 260 articles were captured. 

This web scraping task proved to be the most challenging for AI assistance. While AI tools could generate generic scraping templates, students discovered that these templates could fail on specific NASDAQ pages due to dynamic content loading and anti-bot measures. This failure also became a learning moment: mentors introduced students to browser developer tools for inspecting page structure and to specialized libraries like curl-impersonate for mimicking authentic browser traffic. Students reported that debugging the scraper and iterating between AI and humans as co-mentors in the task gave them deeper hands-on learning experience into how web technologies work.

For each article we employed \texttt{gpt-5-mini} via OpenAI API with constrained generation, forcing the model to output a sentiment score on a fixed scale from -10 to 10, along with the reason for each of the given scores. After scoring, we linked each article to the sectors represented in its associated ETFs. When an article was relevant to multiple ETFs, we recorded the sentiment for each relevant sector. The sentiment scores were then aggregated on a daily basis within each sector: the mean of all articles posted that day produced an average sentiment value per date–sector pair. We did not implement ETF-specific linking in this iteration due to already low sector-level coverage.

Finally, the sentiment data was merged with the ETF price data by date and sector. For days lacking news coverage in a given sector we set the sentiment to 0, indicating neutral sentiment. The resulting dataset contained synchronized price trajectories and contemporaneous sentiment indicators for each sector, and was used for the subsequent price movement modeling. The final coverage between ETF prices and sentiment scores is shown in Table \ref{tab:sentcov}. We acknowledge that this imputation of neutral sentiment may introduce bias into the model, as a lack of news does not necessarily equate to neutrality.

In summary, the students found the sentiment modeling phase the most accessible for AI as a co-mentor. The task for prompting an LLM to score sentiment aligned naturally with their intuition about language, and they iterated on prompt designs with minimal human mentor (scholar) input. However, this accessibility also introduced a risk: students initially trusted the sentiment scores. Mentors used this as an opportunity to discuss the limitations of LLM-based annotation, including inconsistency across similar articles and the lack of ground-truth validation.  

\subsection{Price Movement Modeling}

The price modeling framework is formulated as a two-stage predictive system designed to separately capture the magnitude and the direction of price movements in exchange-traded funds (ETFs). In the first stage, regression-based models are employed to estimate the absolute magnitude of price changes, commonly referred to as price deltas. In the second stage, classification-based models are utilized to predict the directional movement of prices. This modular design allows the framework to disentangle the intensity of market fluctuations from their directional tendencies, thereby providing a more nuanced and interpretable representation of market dynamics.

This two stage architecture was not immediately intuitive to students, who initially proposed predicting raw prices directly. Through a series of stand-up discussions, mentors explained why modeling deltas improves stationarity and why separating magnitude from direction actually simplifies the machine learning problem. The students were then able to use the AI tools to implement both stages, iterating with the mentors, resulting in a cycle that reinforced the conceptual understanding. 

To support large-scale experimentation across multiple ETFs, the pipeline incorporates reliability mechanisms applicable to both stages. Model checkpoints and optimal hyperparameters are persistently stored to enable seamless recovery from interruptions. All model outputs and performance metrics are logged in structured JSON files, and individual test-set predictions are archived in CSV format to support detailed post-hoc analysis. Additionally, automated visualizations are generated for each ETF, overlaying predicted and actual price movements alongside sentiment trends to facilitate comparative analysis and interpretation. We discuss the regression and classification stages now. 

\subsubsection{Regression Stage}
The regression component focuses on predicting absolute price changes (deltas) rather than raw price levels. Modeling price deltas improves the stationarity of the time series (that is, it makes certain statistical properties, like the mean, constant with respect to time), enhances numerical stability, and aligns more closely with the objectives of financial forecasting. To this end, the modeling pipeline evaluates a broad spectrum of approaches, including traditional statistical models, machine learning algorithms, and deep learning architectures, with particular emphasis on quantifying the incremental predictive value of sentiment information.

To transform the time-series data into a supervised learning framework, the original price series is converted into a delta series representing changes over successive time steps. A fixed lookback window (5 days) is then applied to capture temporal dependencies. For non-sequential models such as Support Vector Regression (SVR) and XGBoost, lagged values of both price deltas and sentiment scores are explicitly constructed as input features. In contrast, for sequential models such as Long Short-Term Memory (LSTM) networks, the data is reshaped into three-dimensional tensors of the form (samples times lookback window times features), enabling the model to learn temporal patterns across multiple time steps.

A diverse set of regression models is evaluated to establish a robust performance hierarchy. A five-day moving average serves as a naive baseline to benchmark predictive improvements. Statistical models, including ARIMA and SARIMAX, are employed for univariate and multivariate forecasting, respectively, with sentiment incorporated as an exogenous variable where applicable. Deep learning models based on LSTM architectures are trained under two configurations, using price data alone and using a combination of price and sentiment data, to isolate the contribution of sentiment signals. Additionally, machine learning models such as SVR and XGBoost leverage lagged price and sentiment features to capture nonlinear relationships inherent in financial time series.

To ensure fair and reliable model comparison, all regression models undergo systematic hyperparameter optimization. Grid search procedures are used to tune key parameters, such as network depth and optimizers for LSTM models, and regularization strength and tree depth for SVR and XGBoost. Model validation and hyperparameter grid search were conducted using a walk-forward time-series cross-validation strategy, which prevents look-ahead bias and ensures realistic out-of-sample performance evaluation.

A key objective of the regression framework is to assess the predictive utility of sentiment information. Accordingly, each model is trained in two variants: a univariate version relying solely on historical price deltas, and a multivariate version that augments price information with sentiment scores. Performance differences, evaluated using metrics such as Mean Squared Error (MSE) \cite{kenett_mean_2014}, directly indicate whether the inclusion of sentiment data yields statistically meaningful improvements in predictive accuracy.
%To support large-scale experimentation across multiple ETFs, the regression pipeline incorporates production-level reliability mechanisms. Model checkpoints and optimal hyperparameters are persistently stored to enable seamless recovery from interruptions. 
Furthermore, automated visualizations are generated for each ETF, overlaying predicted and actual price movements alongside sentiment trends to facilitate comparative analysis and interpretation.

As the students leaned on AI tools to help develop the models, they also learned the important lessons of validation, overfitting, and other implementation lessons rapidly. 

\subsubsection{Classification Stage}
While the regression stage estimated the magnitude of price changes, the classification stage is dedicated to predicting the direction of price movement. This component formulates the problem as a binary classification task, reflecting the practical importance of directional forecasts in trading and risk management applications. Price changes are transformed into a binary target variable, where positive price changes are labeled as upward movements and non-positive changes are labeled as downward or neutral movements. This formulation emphasizes directional accuracy, which is often more actionable than precise magnitude estimates in financial decision-making contexts. The predictions from the regression model are then merged with the classifier predictions, where the magnitude of the delta is predicted via the regression model, the sign is predicted via the classifier, and both then multiplied together.

A hierarchical benchmarking strategy evaluated a wide range of classification models. Naive classifiers, such as ``all-up'' and ``all-down'' predictors, are included to diagnose class imbalance and establish lower-bound performance baselines. Classical machine learning models, including Logistic Regression, Support Vector Machines, Decision Trees, Random Forests, and XGBoost, are then evaluated. In addition, deep learning models based on LSTM architectures are trained to exploit sequential dependencies in the data.
LSTM classifiers operate on sliding windows of historical observations. To predict the price direction on a given day, the model processes data from the preceding lookback period, enabling it to learn short-term trends, momentum effects, and temporal patterns that are not captured by static classifiers.

%At the regression stage, two parallel experimental configurations are conducted for each ETF: one relying exclusively on historical price information and one augmenting price features with sentiment scores. This controlled design enables a direct assessment of the incremental value added by sentiment signals. 

Several strategies are adopted to enhance model robustness and generalization. Feature standardization ensures that price and sentiment variables contribute proportionately to the learning process. Hyperparameters are optimized using cross-validated grid search procedures, and model selection prioritizes the F1-score rather than raw accuracy to account for potential class imbalance in financial time series.

By the classification stage, students had developed noticeably stronger debugging instincts. When an XGBoost classifier produced suspiciously high accuracy, students independently hypothesized data leakage before consulting mentors—a marked contrast to their earlier reliance on mentor diagnosis. This progression suggests that repeated exposure to AI-generated errors, coupled with guided debugging, can accelerate the development of critical evaluation skills.

%To manage computational complexity and ensure reproducibility, the classification pipeline incorporates comprehensive progress tracking and result management. All model outputs and performance metrics are persistently logged in structured JSON files, completed model runs are automatically skipped during re-execution, and individual test-set predictions are archived in CSV format to support detailed post-hoc analysis.

\subsection{Results and Model Performance}

\begin{table}[!htp]\centering
\begin{tabular}{llccccc}\toprule
& &\multicolumn{2}{c}{\textbf{w/o Sentiment}} &\multicolumn{2}{c}{\textbf{with Sentiment}} \\\cmidrule{3-6}
\textbf{CM} &\textbf{RM} &\textbf{MSE} &\textbf{MAE} &\textbf{MSE} &\textbf{MAE} \\\midrule
\multirow{5}{*}{Baseline (always down)} &ARIMA &14057.96 &69.80 &12236.56 &63.31 \\
&LSTM &4829.41 &43.72 &5333.01 &45.17 \\
&MA5 &4275.37 &41.95 & & \\
&SVM &4434.59 &37.67 &4698.75 &41.21 \\
&XGBoost &6151.31 &46.90 &6178.09 &47.08 \\
\multirow{5}{*}{Baseline (always up)} &ARIMA &24242.56 &87.57 &21465.47 &81.07 \\
&LSTM &11611.38 &61.51 &12347.92 &62.95 \\
&MA5 &10706.56 &59.75 & & \\
&SVM &10797.27 &55.00 &11158.01 &58.46 \\
&XGBoost &13550.30 &64.30 &13616.42 &64.54 \\
\multirow{5}{*}{Decision Tree} &ARIMA &8575.77 &45.10 &7252.22 &39.70 \\
&LSTM &\ul{2588.23} &27.94 &3205.64 &29.60 \\
&MA5 &\textbf{2301.90} &27.06 & & \\
&SVM &2911.68 &\textbf{25.03} &\textbf{2890.48} &\textbf{26.75} \\
&XGBoost &3562.57 &30.59 &3687.00 &31.47 \\
\multirow{5}{*}{Logistic Regression} &ARIMA &13978.66 &68.82 &12163.43 &62.31 \\
&LSTM &4821.64 &43.33 &5319.63 &44.72 \\
&MA5 &4265.12 &41.49 & & \\
&SVM &4445.98 &37.46 &4663.03 &40.78 \\
&XGBoost &6133.73 &46.45 &6151.83 &46.56 \\
\multirow{5}{*}{LSTM} &ARIMA &10768.45 &58.77 &9746.93 &54.62 \\
&LSTM &4343.57 &38.47 &4676.50 &39.20 \\
&MA5 &3959.76 &37.27 & & \\
&SVM &4279.30 &34.20 &4412.85 &36.99 \\
&XGBoost &5469.65 &41.55 &5469.75 &41.63 \\
\multirow{5}{*}{Random Forest} &ARIMA &9210.49 &47.63 &7566.95 &41.49 \\
&LSTM &3090.49 &30.86 &3440.74 &31.75 \\
&MA5 &2755.15 &29.64 & & \\
&SVM &2966.56 &\ul{26.82} &\ul{3041.15} &\ul{28.70} \\
&XGBoost &4150.48 &33.55 &4237.55 &33.96 \\
\multirow{5}{*}{SVM-RBF} &ARIMA &13176.76 &65.68 &11172.10 &58.80 \\
&LSTM &4611.89 &41.68 &5086.47 &42.98 \\
&MA5 &4058.14 &39.93 & & \\
&SVM &4309.10 &36.12 &4529.77 &39.39 \\
&XGBoost &5933.58 &44.82 &5953.84 &44.95 \\
\multirow{5}{*}{XGBoost} &ARIMA &9432.30 &47.88 &7935.71 &42.25 \\
&LSTM &3017.13 &30.43 &3621.75 &32.05 \\
&MA5 &2706.25 &29.44 & & \\
&SVM &3256.85 &27.34 &3262.03 &29.15 \\
&XGBoost &4166.37 &33.57 &4273.45 &34.35 \\
\bottomrule
\end{tabular}
\caption{Average performance measures for classifier model (\textbf{CM}) and regression model (\textbf{RM}) combinations. The best result is shown in \textbf{bold}, the second best is \underline{underlined}.}\label{tab:mainres}
\end{table}

The overall model performance comparison is shown in Table \ref{tab:mainres}. ARIMA based regressors generally underperformed and achieved the lowest errors when paired with an XGBoost classifier while they incurred the highest errors when combined with baseline classifiers. In contrast, SVM based regressors consistently outperformed other regressors especially when coupled with either decision tree or random forest classifiers suggesting effective capture of temporal dependencies and enhanced predictive stability. LSTM based models also performed well, benefiting from classifiers that model nonlinear or probabilistic decision boundaries such as decision trees yet they remained inferior to SVM regressors overall.

Across all regression models, the baseline classification strategies consistently produced inferior results whereas learning based classifiers, especially decision trees and random forests systematically improved performance, underscoring the importance of directional price movement modeling.

\begin{figure}[htbp]
  \centering
  \includegraphics[width=\linewidth]{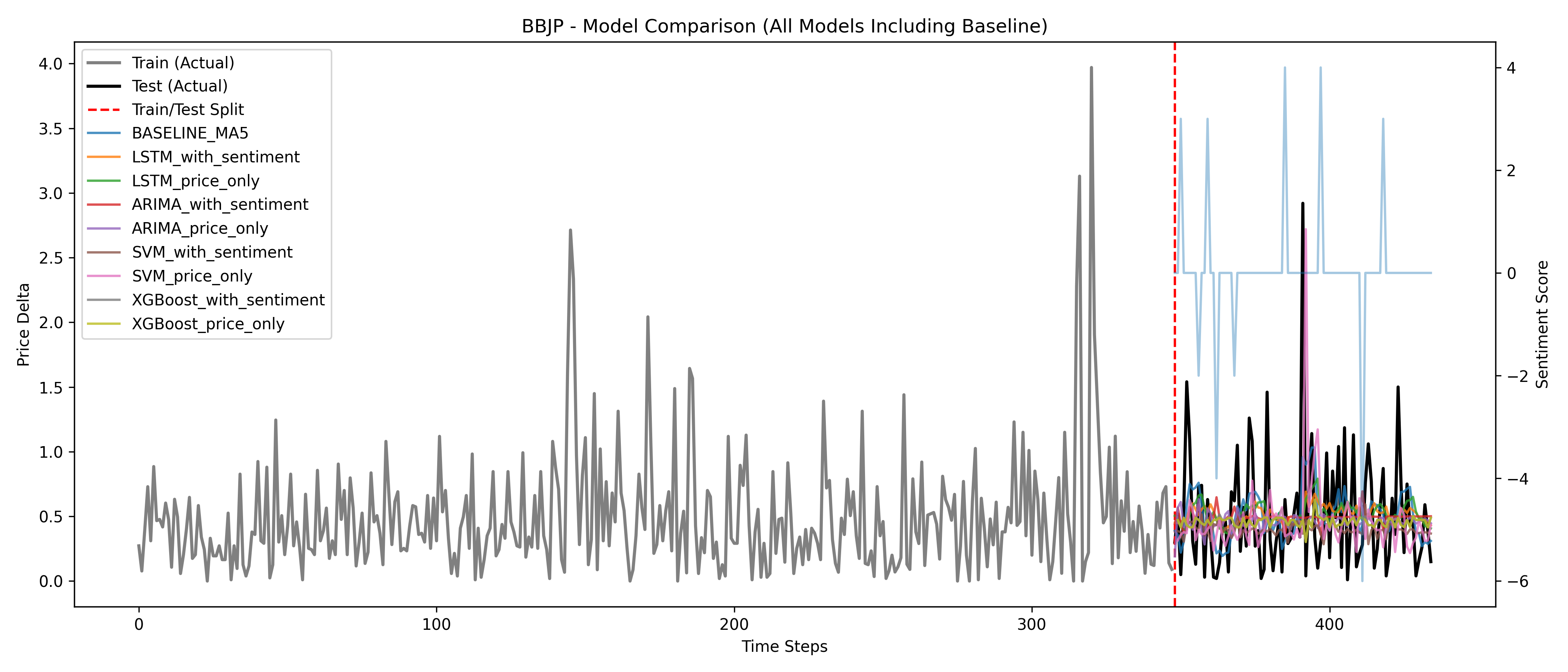}
  \caption{Price delta prediction example (BBJP)}
  \label{fig:predex}
\end{figure}

Additionally, it should be noted that in all cases, the models that included the sentiment as one of their inputs underperformed when compared to their counterparts that had no access to the sentiment information. We hypothesize two explanations: first, sentiment signals may already be incorporated into price movements by the time they are observable in news; second, the explicit sentiment scores introduce noise that degrades model performance. Figure 1 illustrates this behavior: the price-based SVM model captured a large peak representing rapid price change, whereas the sentiment-augmented model did not.

This negative finding, that sentiment hurt rather than helped, was initially disappointing to students, who had invested significant effort in the sentiment pipeline. However, mentors reframed it as a valuable lesson in hypothesis testing: the purpose of the experiment was not to confirm a predetermined outcome but to discover what the data actually revealed. Students reflected that this was one of the most memorable insights of the project, shifting their understanding of research from "proving ideas right" to "finding out what's true."

 The preceding section documented our technical pipeline and its outcomes. We now step back to reflect more broadly on how AI shaped the learning process throughout the project, including its benefits, its failure modes, and the evolving relationship between students, AI-mentors and assistants, and human mentors.

\section{Reflections on AI Use}
\label{sec:reflections}

We present our reflections on using AI across the project, both as a tool for modeling and analysis, and for co-mentorship between AI and humans. AI proved valuable for code organization, debugging, sentiment analysis and as a writing assistant but also introduced maintenance issues, factual errors, and a need for human oversight. We discuss these findings in two subsections that examine AI’s roles as a programming assistant and an analytical tool.

\subsection{AI as a Programming Assistant}

Beyond sentiment generation, AI proved valuable in improving the structural quality and organization of the experimental codebase. In our study, the students have used the ChatGPT chat interface to aid them in writing code and debugging errors. From the students' perspective, AI tools have assisted them in using established programming patterns and best practices. This way, it enhanced code maintainability, reduced redundancy, and improved reproducibility. Additionally, AI support was instrumental in debugging, as it helped identify and correct errors, allowing for faster development. AI further contributed by suggesting efficient pipelines for data pre- and post-processing, enabling easier model comparison and evaluation. Notably, students converged on using the standard ChatGPT interface and primarily utilized its free version, though this occasionally limited access to updated model capabilities. From a data science standpoint, AI was also effective in identifying suitable hyperparameter grids for different classes of models. By proposing reasonable hyperparameter ranges, it helped us tighten the feedback loop which led to faster iteration. At the same time, from the mentors' perspective, the use of AI has allowed us to shift our attention from teaching low-level programming patterns and concepts to explaining and discussing higher-level concepts that were required to build the predictive system.

Despite these advantages, AI exhibited notable limitations, stemming from the fact that it lacked full awareness of the overall research objectives and experimental constraints. As a result, human oversight was required to validate architectural decisions and ensure methodological correctness; AI-generated code occasionally required manual refinement and correction for it to be usable. Moreover, while AI could propose hyperparameters and model structures, it could not independently assess their financial interpretability or empirical suitability without human intervention. Additionally, using AI sometimes resulted in code that was hard to maintain and contained subtle bugs that were difficult for non‑experts to detect. But we also converted all these limitations into learning moments for the students, giving us an opportunity to dig deeper into concepts, offer specific reading exercises, and do real-time debugging together.

From the students' perspective, given that they lacked specialized programming knowledge, it meant that sometimes the behavior of their code changed in an unexpected way, without them having a way to fix the problem without mentor intervention initially as they improved over time. Conversely, from the mentors' perspective, it meant that at certain points in the project, they had to spend additional time on understanding and corrected hard to maintain, AI-generated code. Overall, the debugging sometimes increased the time spent for both, but in retrospect, it also became a deeper engagement experience between scholars and students. 

%These problems required intervention from mentors to resolve. This experience highlighted the need for expert oversight when integrating AI into development workflows.

For example, in programming the news scraper, while AI was able to output generic scaffolding and boilerplate code, the resulting scripts often omitted subtle but essential details, such as the correct code to analyze the source code of the news pages to extract the relevant information. The model also lacked awareness of niche libraries and tools that proved critical in our workflow: for instance, we relied on curl‑impersonate to mimic authentic browser traffic, and the AI we used lacked the knowledge of this particular tool. As a result, the practical usefulness of AI for scraping news stories was limited but offered a good starting point for mentors to engage with students on specific conceptual understanding and doing live coding together. %Most of the core logic of the scraper had to be written and refined manually by developers with relevant experience. 
Similarly to what we said above, this meant that while students did have an opportunity to try their hand at building the scraper, deeper engagement with mentors was ultimately required to being it to a usable state. 

%REMOVED: ... correct code to traverse the DOM tree to extract ...

In summary, AI functioned most effectively as an intelligent programming assistant rather than an autonomous developer. It excelled in tasks related to code organization, sentiment processing, debugging, and documentation, while strategic decisions, including model design, experimental validation, and domain-specific reasoning needed to remain under human oversight and control. This complementary interaction points to a potential use of AI as an augmentation tool in research programming, rather than a replacement for human judgment.

%\subsection{AI as a presentation and writing assistant}

%During manuscript preparation we employed AI extensively for text editing and for proposing strategies that would best convey our findings. The model proved reliable at correcting grammar in short excerpts and could generate first‑draft versions of some sections, which reduced the initial writing effort. Nonetheless, the final paper required intensive human revision to achieve coherence, logical flow, and stylistic consistency. Several failure modes emerged during this process. In a few drafts the AI inserted fabricated analytical steps that had never been performed. Additionally, it also misidentified certain ETFs when describing their sector assignments, producing statements that contradicted our data tables. Each of these inaccuracies demanded manual verification and correction. The experience illustrates that while AI can accelerate drafting and formatting tasks, it cannot yet replace the critical judgment of a human editor in scientific writing. In sum, for presentation and composition tasks we view AI as an aid but not a substitute for thorough human editing. It became an important learning experience for the students to know how and when to even use AI, and how creative and critical thinking, writing, and discernment are critical skills even in the age of AI. 

\subsection{AI as an Analysis Tool}

While AI played a supportive yet non-autonomous role in the programming workflow of this research, its primary and most effective application was in transforming unstructured news statements into structured sentiment scores. This task benefited significantly from AI’s natural language understanding capabilities, as it enabled consistent and scalable sentiment extraction from large volumes of textual data, something that would be hard to achieve using more classical machine learning methods. 

These reflections provide the raw material for addressing our research questions. In the following section, we synthesize our observations into answers to RQ1–RQ3 and draw out implications for educators considering similar approaches.

\section{Discussion}
\label{sec:disc}

We synthesize our findings by returning to the three RQs that motivated the study, and then discuss the broader implications for AI-augmented education. 

\noindent \textbf{RQ1: How does AI-augmented mentorship influence the learning process compared to traditional methods of instruction?} \\

Our experiment suggests that, at least for our specific cohort, LLM-based tools can serve as an effective co-mentor for students with limited domain expertise, but its role is fundamentally different from that of a human mentor. Within the context of our study, AI excelled at providing just-in-time tactical support: generating code scaffolding, suggesting hyperparameter ranges, and offering quick explanations of unfamiliar concepts. This enabled students to engage with the project immediately rather than waiting for prerequisite instruction. 

However, LLM-based tools could not replace the strategic guidance that human mentors provided. Students needed mentors to frame problems, validate architectural decisions, and help them recognize when AI-generated outputs were incorrect. The most effective co-mentoring configuration was rather {\it{complementary}}: AI handled routine implementation tasks, freeing human mentors to focus on higher-level reasoning, debugging and conceptual explanation. 

Overall, AI tools were effective at tactical guidance, as well as helping with the repetitive tasks of software development: they was always available to the students, and were able to read and understand large codebases faster than the human mentor. At the same time, human mentors were required for making larger, strategic decisions: they had domain knowledge and intuition available, coupled with having a complete vision for the overall project.

We also observed an interesting developmental arc. Early in the project, students accepted LLM outputs with little scrutiny; by the end, they had developed skepticism and debugging skills that allowed them to identify errors independently. This progression suggests that working with imperfect AI outputs, albeit under mentor supervision, may itself be a valuable pedagogical intervention.

\noindent \textbf{RQ2: What are the most frequent failure modes encountered when using AI in a project-based learning context?} \\

We identified three recurring categories of AI failure: 
\begin{enumerate}
    \item \textbf{Plausible but incorrect code:} AI-generated code often ran without errors but could contain errors or methodological flaws, including issues associated with assumptions around modeling, feature construction, etc. This is where the expertise of the human mentors became essential, but also where the students could participate in learning through debugging exercise. 
    \item \textbf{Lack of specialized knowledge:} LLM-based tools were great at generating standard templates, but could miss niche libraries that were critical to performing the task at hand. This again required mentors intervention to identify such issues and also walk the students through the nuances of web scraping. 
    \item \textbf{Long term context windows:} As the project lasted over a period of time, it became clear that the awareness of overall research objectives with the AI system became limited, which carried the risk of incorrect advice. We re-oriented ourselves to regular meetings during the analysis and paper writing to ensure alignment. 
\end{enumerate}

In our experience, each of these failure modes actually became a teaching moment, and opportunity for hands-on learning. These are occasions for deeper engagement with underlying concepts. 

\noindent \textbf{RQ3: Do LLM‑generated sentiment scores improve the predictive performance of ETF price forecasting models?} \\

Contrary to our expectations, sentiment features did not improve model performance. We offer two interpretations: 1) sentiment information is already priced into ETF movements by the time it appears in the news articles, rendering explicit sentiment features redundant; and 2) the LLM-generated scores were not completely reliable, warranting additional research.

%In our experiment we found that AI can provide useful tactical guidance for students with little programming or data‑science experience, enabling them to assemble a working modeling pipeline quickly. The system also integrated smoothly into the data‑analysis workflow; it powered the sentiment analysis module that enriched the market data set. However, the model was unable to make correct strategic decisions, and human oversight remained essential for overall research design.

%When we turned to programming tasks, AI proved efficient at accelerating routine operations and implementing common patterns. Yet it lacked familiarity with some of the specialized tools we relied on for web scraping (e.g. curl‑impersonate) and it struggled to generate robust code for news page source code parsing and traversal without expert input. This shortfall meant that developers still had to write and refine core scraping logic manually.

%For writing, AI was helpful in drafting section outlines and initial prose, and it improved the language of small text snippets. Nevertheless, the generated content required further editing before it met publishable standards.

In summary, AI functions best as a powerful augmenting tool for novices in educational settings: it assists with tactical decisions, routine programming tasks, and basic writing support. Beyond these roles, effective use continues to depend on expert or mentor guidance, particularly for higher‑level strategic choices and specialized intuition. The experiment suggests that AI can help students explore diverse programming patterns and make small‑scale decisions, but mentorship remains crucial for ensuring rigorous research outcomes. For educators, this means treating AI as a guided support tool rather than a replacement for mentorship, and explicitly building verification and skepticism into the learning goals. While we started with a project idea on developing machine learning models for predicting ETF movements, this project an important learning experience for the students to know how and when to even use AI, and how creative and critical thinking, writing, and discernment are critical skills even in the age of AI. Simple routines such as regular check-ins and shared documentation also help keep AI-assisted work transparent and allow errors to be caught early.

\section{Limitations and Future Work}
\label{sec:lim}

Our study was carried out with a single cohort of high‑school and early‑undergraduate students from US-style educational institutions over the summer, so caution is warranted when extrapolating these findings to other contexts or institutions. Additionally, the primary objective was to investigate how AI can act as a mentoring companion for novices in research and model performance metrics on the actual modeling were secondary to this goal. As such, predictive accuracy should be viewed as exploratory rather than conclusive evidence of any kind of model performance. Additionally, tt is important to note that the capabilities of LLMs and LLM-based tools evolve rapidly; therefore, our findings are specific to the versions of the tools available during the summer of 2025.

On the forecasting side, results may be affected by sparse and uneven news coverage, sector-level sentiment aggregation that can dilute ETF-specific signals. While we took steps to reduce leakage risk, alternative sentiment extraction methods remain an important direction for future work. News coverage gaps left about one‑tenth of days without sector‑specific sentiment, and we imputed zero sentiment for those dates, potentially biasing the signal. Moreover, LLM-based sentiment scoring is not necessarily robust, introducing noise that could mislead downstream models, and for this work, we did not validate the LLM scores against human-coded ground truth. These limitations point to the importance of contextualizing our results and pursuing further research to validate the educational benefits of AI support.

Future work should broaden the participant base to include more diverse cohorts. Longitudinal studies tracking student learning outcomes over time would also clarify the lasting impact of AI co-mentorship.  

%----------------------------------------------------------------------------------------
%  REFERENCE LIST
%----------------------------------------------------------------------------------------
\vspace{4\baselineskip}\vspace{-\parskip} % Creaters proper 4 blank line spacing.
\footnotesize % Makes bibliography 10 pt font.

\printbibliography

%----------------------------------------------------------------------------------------

\end{document}